\documentclass[conference,10pt]{IEEEtran}
\IEEEoverridecommandlockouts
\usepackage{cite}
\usepackage{amsmath}
\usepackage{amssymb,amsfonts}
\usepackage{algorithmic}
\usepackage{graphicx}
\usepackage{textcomp}
\usepackage[dvipsnames]{xcolor}
\usepackage{soul}
\usepackage{tikz}
\usepackage{transparent}        
\usepackage[ruled,vlined]{algorithm2e}
\usepackage{pgfplots, pgfplotstable}
\usepgfplotslibrary{groupplots}
\usepgfplotslibrary{fillbetween}
\usetikzlibrary{shapes,arrows,calc,spy}
\pgfplotsset{grid style={dashed}}

\def\BibTeX{{\rm B\kern-.05em{\sc i\kern-.025em b}\kern-.08em
    T\kern-.1667em\lower.7ex\hbox{E}\kern-.125emX}}

\definecolor{camel}{rgb}{0.76, 0.6, 0.42}%

\usepackage[normalem]{ulem}

\newenvironment{list4}{
  \begin{list}{$\bullet$}{%
      \setlength{\itemsep}{0.05cm}
         \setlength{\labelsep}{0.2cm}
      \setlength{\labelwidth}{0.3cm}
      \setlength{\parsep}{0in} 
      \setlength{\parskip}{0in}
      \setlength{\partopsep}{0in}
      \setlength{\leftmargin}{0.17in}}}
      {\end{list}}

\newcommand{\E}{\ensuremath{\mathbb{E}}}
\newcommand{\etal}{\textit{et al. }}
\newcommand{\ie}{\text{i.e., }}

\newcommand{\vect}[1]{\boldsymbol{#1}}
\DeclareSymbolFont{matha}{OML}{txmi}{b}{it}
\DeclareMathSymbol{\varv}{\mathord}{matha}{118}

\makeatletter

\newenvironment{shiftedflalign*}{%
    \start@align\tw@\st@rredtrue\m@ne
    \hskip\parindent
}{%
    \endalign
}
\makeatother

\begin{document}
%
%
%
%

\title{
\LARGE Onboard Real-Time Multi-Sensor Pose Estimation for Indoor Quadrotor Navigation with Intermittent Communication}

\author{\IEEEauthorblockN{Loizos Hadjiloizou\IEEEauthorrefmark{2}, Kyriakos M. Deliparaschos\IEEEauthorrefmark{1}, Evagoras Makridis\IEEEauthorrefmark{3} and Themistoklis Charalambous\IEEEauthorrefmark{3}\IEEEauthorrefmark{4}}
\IEEEauthorblockA{\IEEEauthorrefmark{2}KTH Royal Institute of Technology, Stockholm, Sweden, Email: loizosh@kth.se}
\IEEEauthorblockA{\IEEEauthorrefmark{1}Cyprus University of Technology, Limassol, Cyprus, Email: k.deliparaschos@cut.ac.cy}
\IEEEauthorblockA{\IEEEauthorrefmark{3}University of Cyprus, Nicosia, Cyprus, Emails: {surname.name}@ucy.ac.cy}
\IEEEauthorblockA{\IEEEauthorrefmark{4}Aalto University, Espoo, Finland, Email: themistoklis.charalambous@aalto.fi}}


\maketitle

%
%
%
%
\begin{abstract}
We propose a multisensor fusion framework for onboard real-time navigation of a quadrotor in an indoor environment, by integrating sensor readings from an Inertial Measurement Unit (IMU), a camera-based object detection algorithm, and an Ultra-WideBand (UWB) localization system. The sensor readings from the camera-based object detection algorithm and the UWB localization system arrive intermittently, since the measurements are not readily available. We design a Kalman filter that manages intermittent observations in order to handle and fuse the readings and estimate the pose of the quadrotor for tracking a predefined trajectory. The system is implemented via a Hardware-in-the-loop (HIL) simulation  technique, in which the dynamic model of the quadrotor is simulated in an open-source 3D robotics simulator tool, and the whole navigation system is implemented on Artificial Intelligence (AI) enabled edge GPU. The simulation results show that our proposed framework offers low positioning and trajectory errors, while handling intermittent sensor measurements.
\end{abstract}


%
%
%
%
\section{Introduction}

During the past decade, advancements in wireless communications and computer vision for multi-rotor Unmanned Aerial Vehicles (UAVs) have led to the transition towards fully autonomous navigation missions in either outdoor or indoor environments. Missions such as search and rescue, facility monitoring and inspection, and warehouse inventory management, are time- and safety-critical, making reliable information about the UAV's pose necessary. The quadrotor's position in space along with its orientation form the notion of pose. The pose of the quadrotor should be estimated in (almost) real-time and as precisely as possible, as indoor missions  usually require high accuracy robustness, and timely reactions to abrupt changes in the environment.



Although advanced techniques using multi-sensor data fusion exist, they mostly combine cues from Inertial Measurement Unit (IMU) devices and cameras for Simultaneous Localization and Mapping (SLAM).
Such visual-inertial approaches, however, accumulate errors in pose over time due to sensors' noise and modelling errors 
In outdoor environments this problem could be effectively addressed by incorporating global position measurements (e.g. GPS) to advance the pose estimation; see, for example, \cite{2018:Chli} and references therein. In indoor environments, on the other hand, it is easier to have a pre-existing mapping and features with known locations. 

Vision-based approaches  use off-board and on-board visual sensing to localize UAVs. Off-board sensing relies on expensive fixed motion-capture systems with high frame-rate off-board cameras \cite{preiss2017crazyswarm,sa2018dynamic},  rendering such methods, in regards to portability, too cumbersome to use. In contrast, on-board sensing uses on-board monocular~\cite{lim2015monocular,s20030919} and stereo~\cite{fraundorfer2012vision} cameras, offering a more practical and cheaper solution, although their limited performance under illumination and viewpoint changes  \cite{maffra2019real}.
%
Wireless-based approaches and, more specifically, Ultra-Wideband (UWB) wireless technology, has attracted interest from both researchers and practitioners for UAV localization tasks \cite{papastratis2018indoor, makridis2020towards}, in the absence of GPS signals,  mainly due to  system's large scalability, low cost and ease of installation.  However, UWB-based localization approaches can be proved unreliable in cases there is no line of sight, resulting in noisy measurements with large variance and communication delays.
Several approaches for fusing data from UWB localization systems and SLAM have been proposed in \cite{tiemann2018enhanced, yang2021uvip, nguyen2021range} to provide a more accurate drift-reduced estimate of robots' position in indoor environments. The authors in \cite{yang2021uvip}, however, were the first to propose such a multi-sensor fusion approach with possible sensor failures. Their proposed positioning system is based on an optimization-based sensor fusion, independent of the quadrotor state model, that can handle sensor failures, while improving the positioning accuracy and robustness.

In this work, we target the pose estimation problem with a  multisensor fusion framework, by fusing IMU measurements with UWB technology and the YOLO object detection algorithm firstly proposed by Redmon \etal in \cite{redmon2016you}. 
Combining these methods with the intermittent sensor measurements, we propose a reliable positioning system for indoor quadrotor navigation. Our contributions are as follows:
\begin{list4}
\item We introduce a robust hybrid multisensor data fusion framework  utilizing a Kalman filter to fuse IMU data and possibly intermittent and noisy measurements from YOLO and UWB. 
\item We provide a framework for HIL simulation of autonomous on-board indoor navigation to demonstrate how  pose sensing and estimation respond in real-time, facilitating the transition to real hardware implementation.
\end{list4}

The rest of this paper is structured as follows. 
Section~\ref{sec:system_description} defines the coordinate systems, and presents the derivation of the quadrotor model. Section~\ref{sec:sensors} presents the sensing technologies considered for quadrotor localization. Section~\ref{sec:multi_sensor_fusion} introduces the multisensor fusion with intermittent measurements framework and the controller adopted in this work. The evaluation of the developed framework and its performance are described in Section~\ref{sec:experimental_validation}. Lastly, conclusions and directions for future work are given in Section \ref{sec:conclusions}.

%
%
%
%
\textbf{Notation.} In this paper we denote vectors, matrices and sets by bold lowercase, uppercase and calligraphic uppercase letters, respectively. Sets $\mathbb{R}$, $\mathbb{R}_{+}$, and $\mathbb{N}$ represent real, nonnegative real and natural numbers, respectively. $I_{p\times p}$ denotes the identity matrix of dimension $p$, $p\in{\mathbb N}$, unless its dimensions are obvious, in which case it is represented by $I$. For any  matrix $A\in \mathbb{R}^{p\times m}, (p,m)\in {\mathbb N}\times {\mathbb N}$, we denote its transpose by $A^{T}$, and for $m=p$, we denote its inverse by $A^{-1}$. Let $\mathrm{diag}\{A\}$ be the matrix with zero elements everywhere but along the diagonal entries $A_{ii},~i=1,\ldots,p$. Let $A\succeq 0$ and $A\succ 0$ be a positive semi-definite and a positive definite matrix $A\in \mathbb{R}^{p\times p}$, respectively. We denote the expectation of $\cdot$ by $\mathbb{E}\{\cdot \}$. 
For brevity, we denote the sine and cosine of an angle $\theta$ by $s\theta \equiv \sin(\theta)$ and $c\theta \equiv \cos(\theta)$, respectively.

%
%
%
%
\section{System Description}\label{sec:system_description}
\subsection{Coordinate Systems}
\begin{figure}[h]
	\begin{center}
		\includegraphics[width=0.95\columnwidth]{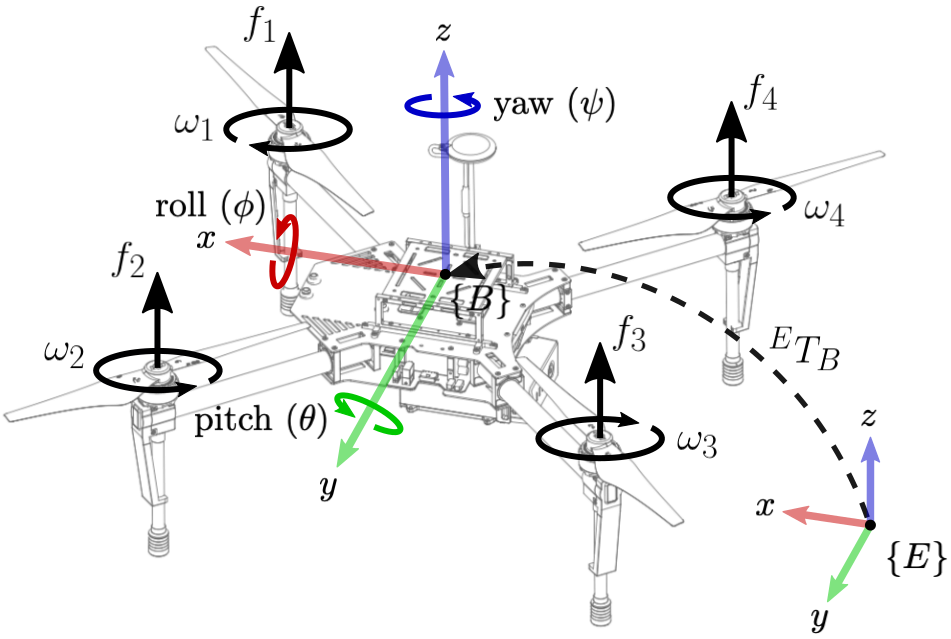}
		\caption{DJI M100 quadrotor model with coordinate frame.}
		\label{fig:quadrotor}
	\end{center}
\end{figure}

We determine two coordinate frames using the standard right-handed robotics convention (see~Fig.\ref{fig:quadrotor}). The Earth's inertial frame $\{E\}$ (right-hand rule) follows the East-North-Up (ENU) reference system, where $+x$ axis points to the east, $+y$ to the north and $+z$ points upwards. The quadrotor's Body frame $\{B\}$ is coincident to the quadrotor's center of mass representing the absolute position of the quadrotor. It follows the Forward-Left-Up (FLU) reference system, where $+x$ gives forward horizontal movement, $+y$ gives left horizontal movement and $+z$ gives up vertical movement among their axes.

\subsection{Quadrotor Model}\label{sec:quadrotor_dynamics}
Quadrotors are by nature unstable non-linear complex systems made up, as the name implies, of four rotors. Each rotor consists of a propeller and a motor producing an angular velocity $\omega_i$, which implies a thrust force $f_i$, where $i$ corresponds to the number of the motor as shown in Fig.\ref{fig:quadrotor}. Two of these rotors spin one way, and two of them spin in the opposite direction to prevent unwanted yaw ($\psi$) rotation of the quadrotor's body (\ie conservation of angular momentum). The angular velocities of each rotor correspond to certain rotational coordinates (\ie $\vect{\eta}=(\phi,\theta,\psi)\in\mathbb{R}^3$) and move the quadrotor to different translational coordinates (\ie $\vect{\xi}=(x,y,z)\in\mathbb{R}^3$) in Earth inertial frame $\{E\}$. Euler angles $\phi$, $\theta$ and $\psi$ define the orientation of the quadrotor: $\phi$ (roll) is the angle about the $x$-axis, $\theta$ (pitch) is the angle about the $y$-axis, and $\psi$ (yaw) is the angle about the $z$ axis. The translational coordinates $x$, $y$ and $z$ denote the center of mass of the quadrotor with respect to Earth inertial frame.


The translational equations of motion in Earth frame (expressed by the Newton-Euler formalism) \cite{kendoul2006nonlinear,voos2009nonlinear}, and the rotational equations of motion \cite{sabatino2015quadrotor} (with relatively small quadrotor movement angles) are given by:
\begin{subequations}\label{eq:eom}
\begin{align}
    \ddot{x} &= \frac{f_T}{m} \left(c\phi s\theta c\psi + s\psi s\phi \right), \quad     \ddot{\phi}=\frac{I_{y}-I_{z}}{I_{x}} \dot{\theta} \dot{\psi}+\frac{\tau_{x}}{I_{x}},\nonumber\\
	\ddot{y} &= \frac{f_T}{m} \left(c\phi s\theta s\psi - c\psi s\phi \right),\quad     \ddot{\theta}=\frac{I_{z}-I_{x}}{I_{y}} \dot{\phi} \dot{\psi}+\frac{\tau_{y}}{I_{y}},\nonumber\\
	\ddot{z} &= \frac{f_T}{m} \left(c\phi c\theta \right) - g, \quad\;\;\; \quad\quad    \ddot{\psi}=\frac{I_{x}-I_{y}}{I_{z}} \dot{\phi} \dot{\theta}+\frac{\tau_{z}}{I_{z}},\nonumber
\end{align}
\end{subequations}
where $m$ denotes the mass of the quadrotor, $f_T$ denotes the total thrust force, and $g$ the acceleration of the gravity. The terms $\tau_\theta$, $\tau_\phi$ and $\tau_\psi$ represent the pitch torque, roll torque and yaw torque, respectively, as a function of the angular velocities of the rotors $\Omega_i$ \cite{bouabdallah2007design}.  In addition, $I_x$, $I_y$, and $I_z$, express the moments of inertia of the quadrotor's symmetric rigid body about its three axes.
Since we assume relatively low navigation speed, the angles of rotation are also small, hence we may use the small-angle approximation to simplify and linearize the nonlinear model around the hovering equilibrium point or $\dot{\phi},\dot{\theta},\dot{\psi}\simeq 0$; $s\phi \simeq \phi,~s\theta \simeq \theta,~s\psi \simeq \psi$; $c\phi = 1$; $c\theta = 1$; $c\psi = 1$. First we define the system state vector (\ie $\vect{x}=[x~y~z~\phi~\theta~\psi~\dot{x}~\dot{y}~\dot{z}~\dot{\phi}~\dot{\theta}~\dot{\psi}]^T$). Then, we determine the hovering equilibrium point, $\vect{\bar{x}}=[\bar{x}~\bar{y}~\bar{z}~0~0~0~0~0~0~0~0~0]^T$, reached when using a constant control input $f_T=mg$ reflecting the total thrust when hovering at an arbitrary position $(\bar{x},~\bar{y},~\bar{z})$. 
The resulting linearized system is described by
\begin{align}
\ddot{x}&=g \theta,\quad \ddot{y}=-g \phi, \quad \ddot{z}= \frac{f_{T}}{m},\nonumber\\
\ddot{\phi}&=\frac{\tau_{x}}{I_{x}},\;\;\; \ddot{\theta}=\frac{\tau_{y}}{I_{y}}, \quad\;\; \ddot{\psi}=\frac{\tau_{z}}{I_{z}}.
\end{align}



The continuous-time state-space model of the quadrotor and the sensors measurements are given by
\begin{subequations}
\begin{align}
\mathrm{d}\vect{{x}}(t) &= A \vect{x}(t)\mathrm{d}t + B \vect{u}(t) \mathrm{d}t + \mathrm{d}\vect{w}(t),\\
\mathrm{d}\vect{y}(t) &= C\vect{x}(t)\mathrm{d}t+\mathrm{d}\vect{v}(t),
\end{align}  
\end{subequations}
where $\vect{x} \in\mathbb{R}^{n}$ is the system state vector with $n=12$ states (\ie $\vect{x}=[x~y~z~\phi~\theta~\psi~\dot{x}~\dot{y}~\dot{z}~\dot{\phi}~\dot{\theta}~\dot{\psi}]^T \in\mathbb{R}^{n}$), $\vect{u} \in\mathbb{R}^{m}$ is the control input vector with $m=4$ controls (\ie $\vect{u}=[f_{T}~\tau_{x}~\tau_{y}~\tau_{z}]^{T}$), $\vect{y} \in\mathbb{R}^{p}$ is the measurement vector with $p=9$ sensor measurements (\ie $\vect{y}=[x_{\mathrm{uwb}}~y_{\mathrm{uwb}}~z_{\mathrm{uwb}}~x_{\mathrm{yolo}}~y_{\mathrm{yolo}}~z_{\mathrm{yolo}}~\phi_{\mathrm{imu}}~\theta_{\mathrm{imu}}~\psi_{\mathrm{imu}}]^T$). Let $\vect{w}\in\mathbb{R}^{n}$ and $\vect{v}\in\mathbb{R}^{p}$ be the zero-mean disturbance stochastic processes, representing the process and measurement noise levels. The discrete-time equivalent state-space model of the quadrotor is
\begin{subequations}\label{eq:lti_system}
\begin{align}
\vect{x}_{k+1} &= \Phi \vect{x}_k + \Gamma \vect{u}_k + \vect{w}_k,\label{eq:lti_system-states} \\
\vect{y}_k &= C\vect{x}_k + \vect{v}_k,\label{eq:lti_system-states-measurements}
\end{align}    
\end{subequations}
\noindent where $\Phi=e^{Ah}$ is the discrete equivalent system matrix, $\Gamma=\int_{s=0}^{h} e^{As} B ds$ is the control input matrix, and $h$ is the sampling period (\ie control loop period). Let measurement matrix $C \in \mathbb{R}^{p\times n}$ denote the relationship between the system states and the system output that is to be measured by the sensors. Vectors $\vect{w}_k$ and $\vect{v}_k$ represent the process and measurement noise, assumed as white Gaussian random sequences with zero mean, with $\E\{\vect{w}_k\}=0$, $\E\{\vect{v}_k\}=0$, $\E\{\vect{w} \vect{w}^T\} = W\succeq0$, and $\E\{\vect{v} \vect{v}^T\} = V\succ0$. We denote the \emph{a priori} and \emph{a posteriori} state estimates by $\hat{\vect{x}}_{k|k-1}$ and $\hat{\vect{x}}_{k|k}$, respectively, and we define the corresponding error covariance matrices by
\begin{align}
P_{k|k-1} &\triangleq \E \{(\vect{x}_k - \hat{\vect{x}}_{k|k-1})(\vect{x}_k - \hat{\vect{x}}_{k|k-1})^T\} ,\\
P_{k|k} &\triangleq \E \{(\vect{x}_k - \hat{\vect{x}}_{k|k})(\vect{x}_k - \hat{\vect{x}}_{k|k})^T\}.
\end{align}

\section{Pose Sensing}\label{sec:sensors}

\subsection{Inertial Measurement Unit}
IMU (Inertial Measurement Unit) is an electronic device that measures a body's orientation, velocity, and gravitational forces by combining accelerometer, gyroscope, and magnetometer sensors into one.  It is usually fused with different vision or wireless-based sensors to complement the pose estimation accuracy, or even to act as the primary sensor in cases that other sensors are unavailable. The IMU provides measurements of the quadrotor's orientation in the form of the three Euler angles; roll, pitch, and yaw.

\subsection{UWB-Localization}
UWB is a wireless communication technology which enables the transmission of short pulses with low energy over high bandwidth. This makes UWB highly resistant to multipath interference. In addition, the nature of UWB facilitates precise measuring of Time-of-Flight (ToF) which means that distances can be estimated with high precision. These two features made UWB a popular technology for localization in indoor environments. Localization via UWB signals relies on several wireless transmitters (\ie anchors) that are placed at known locations, and an UWB receiver (\ie tag) placed on the UAV which receives and logs the arrival times of the UWB signals to calculate the UAV's position in space\footnote{The location of the anchors can be also inferred, given that they are in a certain formation and the initial position of the tag is known.}.
\subsection{YOLO-Localization}


Object detection aims to localize different objects in space and assign to them labels regarding their class of object. 
In brief, the YOLO algorithm divides a given image into several grids of certain dimensions where each grid aims at detecting an object based on a number of bounding boxes that are predicted from the grid cells. Each of these bounding boxes is comprised of five elements that measure the coordinates of the object in the image (\ie $x$, $y$), the dimensions of the object in the image (\ie length, width) and a confidence level regarding the accuracy of the bounding box. 
Given the image dimensions, focal length of the camera, pixels on which the landmarks were detected, and actual position of the landmarks in the 3D world, it is possible to estimate the position of the camera (and consequently the quadrotor's position) using trilateration \cite[Eq.(5)]{zhou2019trilateration}. 
\vspace{3pt}\section{Pose Estimation and Control}\label{sec:multi_sensor_fusion}
\vspace{-5pt}
Receiving intermittent measurements from different sensors (\ie the IMU, the UWB, and the camera) can adversely affect the control performance or even destabilize the system, due to delayed and/or dropped measurement packets from the UWB localization system, and failure of the YOLO algorithm to detect enough objects for calculating the pose of the quadrotor. Inspiring from the seminal work \cite{Schenato:2007}, on the LQG problem with independent and identically distributed (i.i.d.) packet dropping communication links, we present an LQG-Servo controller based on a multi-sensor fusion algorithm with intermittent observations, for pose estimation and trajectory tracking.
The problem of designing an optimal feedback controller for such a stochastic system \eqref{eq:lti_system}, can be broken down into two separate problems (due to separation principle of linear systems driven by additive white Gaussian noise): i) the design of an optimal estimator, \ie a Linear Quadratic Estimator (LQE) which estimates the pose of the quadrotor, and ii) the design of an optimal deterministic feedback controller, \ie a Linear Quadratic Servo Controller (LQ-Servo) which drives the states to a certain reference point or trajectory. If the system is observable and controllable, the standard Linear Quatratic Gaussian Servo (LQG-Servo) controller is obtained. 


\hspace{-1cm}
\begin{figure*}[t]
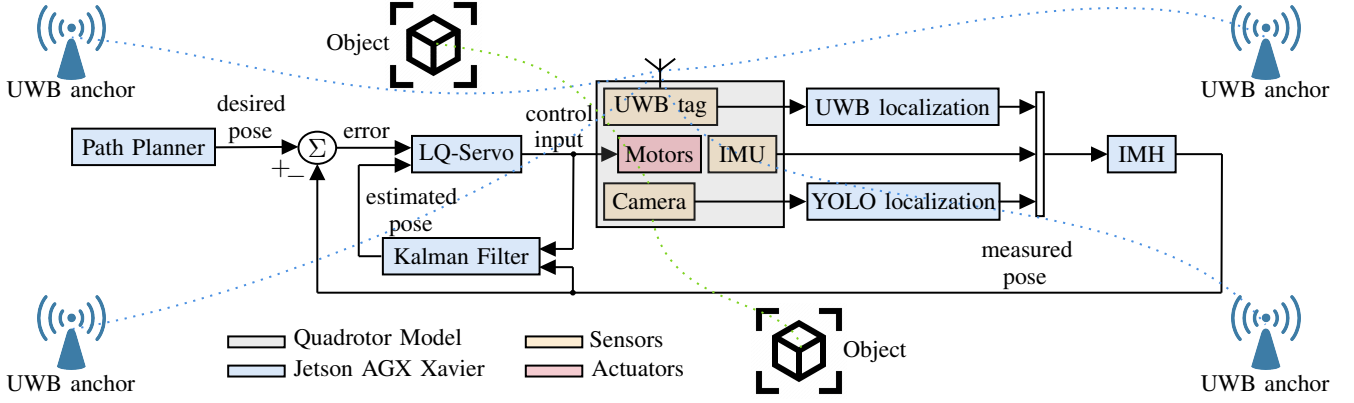

\begin{center}
    \include{figures/control_system_block_diagram}
    \vspace{-15pt}
    \caption{Block diagram of the feedback loop system, showing how the HIL (Jetson AGX Xavier) interacts with the simulation environment (Gazebo).}
    \label{fig:system}
\end{center}
\end{figure*}

\vspace{-5pt}
\subsection{Linear Quadratic Servo (LQ-Servo) Control}\label{sec:lqr}
To follow a reference signal $\vect{r}_k = [r_x,r_y,r_z]^T$ that is generated by the path planner, we augment the state-space model in \eqref{eq:lti_system} by adding an integral state vector, $\vect{i}_{k+1} = \vect{i}_k + \vect{r}_k - E \vect{y}_{k}=-EC \vect{x}_k + \vect{i}_k + \vect{r}_k - E\vect{v}_k$, where $E \in \mathbb{R}^{3 \times {9}}$ is the matrix that selects the controlled states (\ie $E = [\Delta_{1, k}, ( \sim \Delta_{1, k} \wedge (\Delta_{2, k}) ), 0_{3\times3}]$), with $(\sim)$ and $(\wedge)$ being the element-wise logical NOT and AND operators, respectively. Thus, the augmented state-space model is
{\small 
\begin{align}
 \underbrace{\begin{bmatrix} \vect{x}_{k+1} \\ \vect{i}_{k+1} \end{bmatrix}}_\text{{\small $\vect{\bar{x}}_{k+1}$}}
 \!&=\!
  \underbrace{
  \begin{bmatrix}
	\Phi & 0 \\
	-EC & I
   \end{bmatrix}}_\text{{\small $\bar{\Phi}$}}\!
   \underbrace{\begin{bmatrix}
	\vect{x}_{k} \\ \vect{i}_{k} 
   \end{bmatrix}}_\text{{\small $\vect{\bar{x}}_{k}$}}\!
    +\!
   \underbrace{\begin{bmatrix}
	\Gamma \\
	0
   \end{bmatrix}}_\text{{\small $\bar{\Gamma}$}}\!
\vect{u}_k +\!
   \underbrace{\begin{bmatrix}\!
	I & 0\\
	0 & -E
   \end{bmatrix}}_\text{{\small $\bar{E}$}}\!
   \underbrace{\begin{bmatrix}\!
	\vect{w}_k \\
	\vect{v}_k
   \end{bmatrix}}_\text{{\small $\bar{\varv}_k$}} +\!
    \underbrace{\begin{bmatrix}
	0 \\ I
   \end{bmatrix}}_\text{{\small $\bar{I}$}}\!
\vect{r}_k,\nonumber\\
\vect{\bar{y}}_k &= \underbrace{\begin{bmatrix}
    C & 0
\end{bmatrix}}_\text{{\small $\bar{C}$}}
\underbrace{\begin{bmatrix}
    \vect{x}_k \\ \vect{i}_k
\end{bmatrix}}_\text{{\small $\bar{\vect{x}}_k$}}\!+\: \vect{v}_k,\nonumber
\end{align}}%
where $\bar{\varv}_k$ and $\vect{v}_k$ are discrete-time Gaussian white noise processes with zero-mean value and {covariances} $\mathbb{E}\left\{\bar{\varv}_k \bar{\varv}_k^{T}\right\}=V_{1}$, $\mathbb{E}\left\{\bar{\varv}_k \vect{v}_k^{T}\right\}=V_{12}$, and $\mathbb{E}\left\{\vect{v}_k \vect{v}_k^{T}\right\}=V_{2}$,
\begin{align}
\mathbb{E}\left\{\left[\begin{array}{l}
    \bar{\varv}_k \\
    \vect{v}_k
    \end{array}\right]\left[\begin{array}{l}
    \bar{\varv}_k \\
    \vect{v}_k
    \end{array}\right]^{T}\right\}=\left[\begin{array}{cc}
    V_{1} & V_{12} \\
    V_{12}^{T} & V_{2}
    \end{array}\right].\nonumber
\end{align}
However, since $\bar{\varv}_k$, and $\vect{v}_k$ are correlated, then 
\begin{align}
    \mathbb{E}\left\{\bar{\varv}_k \vect{v}_k^{T}\right\}=V_{12}=\mathbb{E}\left[\begin{array}{ll}
    \vect{w}_{k} \vect{v}_{k}^{T} & \vect{v}_{k} \vect{v}_{k}^{T}
    \end{array}\right]=\left[0 \quad V_{2}\right].\nonumber
\end{align}
{As long as the availability of the measurements are independent of the control inputs, which is the case in this work,  the certainty equivalence principle holds. Therefore, t}he optimal control law $\vect{u}_k^* = -L\vect{\bar{x}}_{k} = -L^{\hat{x}} \vect{\hat{x}}_{k|k-1} - L^{i} \vect{i}_k$ can be found by minimizing the linear quadratic criterion in \eqref{eq:lq_cost}
\begin{align}
J &= \mathbb{E}\left[\vect{\bar{x}}_{N}^{\mathrm{T}} \bar{Q}_N \vect{\bar{x}}_{N}+\sum_{k=0}^{N-1}\left(\vect{\bar{x}}_{k}^{\mathrm{T}} \bar{Q}_{k} \vect{\bar{x}}_{k}+\vect{u}_{k}^{\mathrm{T}} R_{k} \vect{u}_{k}\right)\right],\label{eq:lq_cost}
\end{align}
where $\bar{Q}_N\succeq 0$, $\bar{Q}_{k}\succeq 0$ are the final and stage state error weighting matrices, respectively, and $R_{k}\succ0$ is the stage control weighting matrix for the LQ problem. The optimal control gain $L_k = \begin{bmatrix} L_k^{\hat{x}} & L_k^{i}\end{bmatrix}$ is the standard state-feedback controller gain given by
\begin{align}
L_{k}=\left(\bar{\Gamma}^{T} S_{k+1} \bar{\Gamma}+R_k\right)^{-1} \bar{\Gamma}^{T} S_{k+1} \bar{\Phi}
\end{align}
and where $S_{k}$ satisfies the  discrete-time algebraic Riccati equation (DARE) 
\begin{align}\label{eq:DARE}
S_{k} &= \bar{Q}_k + \bar{\Phi}^{T} S_{k+1} \bar{\Phi} \nonumber\\
&\quad-\bar{\Phi}^{T} S_{k+1} \bar{\Gamma}\left(\bar{\Gamma}^{T} S_{k+1} \bar{\Gamma} + R_k\right)^{-1} \bar{\Gamma}^{T} S_{k+1} \bar{\Phi} . 
\end{align}
By considering the infinite horizon problem with $\bar{Q}_k = \bar{Q}$ and $R_k = R$ for all time steps $k$, and based on the assumption that the pairs $(\bar{\Phi},\bar{\Gamma})$ and $(\bar{\Phi},\bar{Q}_k^{1/2})$ are controllable and observable, respectively, the positive semi-definite solution of \eqref{eq:DARE} always exists \cite{Chen:1995}. Then, the controller $L_{\infty}$ becomes
\begin{align}
L_{\infty} = \left(\bar{\Gamma}^{T} S_{\infty} \bar{\Gamma}+R\right)^{-1} \bar{\Gamma}^{T} S_{\infty} \bar{\Phi} ,
\end{align}
where $S_{\infty}$ is the positive semi-definite solution of the DARE,
\begin{align} \label{eq:DARE-infty}
S_{\infty} &= \bar{Q} + \bar{\Phi}^{T} S_{\infty} \bar{\Phi} -\bar{\Phi}^{T} S_{\infty} \bar{\Gamma}\left(\bar{\Gamma}^{T} S_{\infty} \bar{\Gamma} + R\right)^{-1} \bar{\Gamma}^{T} S_{\infty} \bar{\Phi}.\nonumber
\end{align}


\vspace{2pt}
\subsection{Intermittent Measurements Handler (IMH)}
\vspace{-3pt}
To handle the propagation of the intermittent observations to the Kalman filter, we introduce $H_k = \Delta_{k}\bar{C}$, denoting the measurement matrix that enables the observations of the system dynamics through the sensors $j=\{1,2,3\}$ at time instance $k$. Matrix $\Delta_{k} \in\mathbb{R}^{p\times p}$ is a block diagonal matrix consisting of three submatrices/blocks that determine whether observations from sensor $j$, are available at time instance $k$,
\begin{align}
	\Delta_k=\left[\begin{array}{lll}
    \Delta_{1,k} & 0_{3\times 3} & 0_{3\times 3} \\
    0_{3\times 3} & \Delta_{2,k} & 0_{3\times 3} \\
    0_{3\times 3} & 0_{3\times 3} & \Delta_{3,k} \\
    \end{array}\right],
\end{align}
where block $\Delta_{j,k}$ indicates that sensor $j$ collects measurements at time instance $k$. That is, if measurements are available, then $\delta^j_k=1$; otherwise,  $\delta^j_k=0$. Hence, the arrival of observation from sensor $j$ at time $k$ can be expressed as 
\begin{flalign}
	    \Delta_{j,k} =
            \begin{cases}
            I_{3\times3}, & \text{if measurement avail., $ \delta^j_k = 1 $,}\\
            0_{3\times3}, & \text{if measurement not avail., $ \delta^j_k = 0$.}
            \end{cases} &&
\end{flalign}

\subsection{Kalman Filter for Multisensor Fusion}
\vspace{-3pt}
Since we approximate the dynamics of the quadrotor by discrete-time linear equations, and it is assumed to have Gaussian process and measurement noise, we propose the use of a Kalman filter, which is the optimal estimator under the aforementioned assumptions. Nevertheless, we replace the static measurement matrix $\bar{C}$ which is used in classical filtering approaches, with a dynamic measurement matrix $H_{k}$ to handle the intermittent observations from the sensors \cite{1333199}. The Kalman filter essentially determines a set of observer gains $K_k$ to minimize the estimation error covariance. For the estimation error, $\tilde{\vect{x}}_{k+1}\triangleq \bar{\vect{x}}_{k+1}-\hat{\vect{x}}_{k+1|k}$, we have
\begin{align}
\tilde{\vect{x}}_{k+1}=\bar{K}_k \left(\left[\begin{array}{l}
\bar{\Phi} \\
H_k
\end{array}\right] \tilde{\vect{x}}_{k}+\left[\begin{array}{ll}
\bar{E} & 0 \\
0 & I
\end{array}\right]\left[\begin{array}{l}
\bar{\varv}_{k} \\
\vect{v}_{k}
\end{array}\right]\right)+\bar{I} \vect{r}_{k}, \nonumber
\end{align}
where $\bar{K}_k=\left[I \quad -K_k\right]$. The error covariance of the augmented state-space model is
\begin{align}
P_{k+1} &=\mathbb{E}\left\{\tilde{x}_{k+1}{\tilde{x}}_{k+1}{ }^{T}\right\} \nonumber \\
&=\bar{K}_k\left[\begin{array}{ll}
\bar{\Phi} P_{k} \bar{\Phi}^{T}+\bar{E} V_{1} \bar{E}^{T} & \bar{\Phi} P_{k} H_k^{T}+\bar{E} V_{12} \\
H_k P_{k} \bar{\Phi}^{T}+V_{12} \bar{E}^{T} & H_k P_{k} H_k^{T}+V_{2}
\end{array}\right] \bar{K}_k^T. \nonumber
\end{align}
Hence, the Kalman filter for the augmented state-space model with intermittent observations and correlated process-measurement noise becomes
\begin{subequations}
	\begin{flalign}
    K_{k} &= \left(\bar{\Phi} P_{k} H_k^{T}+\bar{E} V_{12}\right) \left(H_k P_{k} H_k^{T}+V_{2}\right)^{-1},\nonumber\\
    P_{k+1} &=\bar{\Phi} P_{k} \bar{\Phi}^{T}+\bar{E} V_{1} \bar{E}^{T} - \left(\Phi P_{k} H_k^{T}+\bar{E} V_{12}\right)\nonumber\\
    &\quad\quad \left(H_k P_{k} H_k^{T}+V_{2}\right)^{-1}\left(H_k P_{K} \bar{\Phi}^{T}+V_{12} \bar{E}^{T}\right),\nonumber\\
    \hat{\vect{x}}_{k+1|k} &= \bar{\Phi} \hat{\vect{x}}_{k|k-1} +\bar{\Gamma} \vect{u}_{k}+K_{k}\left(\bar{\vect{y}}_{k}-{H}_k \hat{\vect{x}}_{k|k-1}\right).\nonumber
	\end{flalign}%
\end{subequations}



%
%
%
%
\vspace{3pt}
\section{Experimental Validation}\label{sec:experimental_validation}

\subsection{Experimental setup}\label{subsec:experimental_setup}
In this work, we simulate the 
proposed system in Gazebo open-source 3D robotics simulator tool. We model the reference indoor environment used for simulation and measurements using several UWB tags and objects (detectable and non-detectable) by the YOLO v3 algorithm \cite{redmon2018yolov3}. Subsequently, we implement the control system as a Robot Operating System (ROS) node in C{/}C++, Python, and utilize an NVIDIA Xavier AI embedded module for all required computations\footnote{The base project code presented in this work, called ORTILo (Onboard Real-Time Indoor Localization), is hosted in Atlassian Bitbucket (https://bitbucket.org/LoizosHadjiloizou/ortilo) as a Git repository.}. The communication link between the quadrotor, sensors, estimation, and control operates through ROS messages exchange. The validation of the proposed multisensor fusion localization involves a HIL simulation of an embedded control system (see Fig.\ref{fig:system}), where the dynamical model of the quadrotor interfaces with the actual controller via a communication link. As shown in Fig.\ref{fig:system}, the IMU senses the orientation of the quadrotor and feeds the measurements to the estimator.
At the same time, the UWB anchor receives ranging signals transmitted by the tags to provide the UWB localization node with the calculated distances, which in turn estimates the position using multilateration, and inputs the measured position to the estimator. Camera captured images of the indoor environment input the YOLO localization component 
after which it scans them for objects included in the Microsoft Common Objects in COntext (MS COCO) dataset\cite{lin2015microsoft} and publishes a ROS message with metadata for every bounding box containing a detected object. The published metadata are then parsed and the YOLO algorithm executes to calculate the position of the quadrotor, and send the measurements to the estimator.

\subsection{Experimental results}\label{subsec:experimental_results}
The target of the quadrotor in the following experiments is to follow a given trajectory while maintaining a constant altitude. The process and measurement covariance matrices of the Kalman filter are set for all experiments to $ W = I_{12\times12} $ and $V = \mathrm{diag}(0.05I_{3\times3}, 0.08I_{3\times3},0.01I_{3\times3})$, respectively.
We evaluate the system performance through the mean square errors (MSEs) of: \emph{i)} the state estimation, and \emph{ii)} path following. Specifically, we use the mean MSE  alongside the 25-th, the 50-th (median MSE), and 75-th percentiles\footnote{k-th percentile is a value below which a given percentage $k$ of values in its frequency distribution falls.}. The difference between the quadrotor actual position (obtained by the simulator) and its corresponding estimated position determines the MSE of the state estimation, thus providing a good indication of the pose estimation performance when sensors measurements arrive intermittently. The MSE of the path following represents the deviation between the desired (planned) path and the actual trajectory the quadrotor follows. We collect data from 20 similar tests to compensate for the stochastic aspects of the experiments.

Table~\ref{table:estimation_error} shows the MSE of the state estimation when only the IMU and UWB sensors are used (scenario 1), and when all three sensors are used (scenario 2). 
Scenario 2 results (regarding x-axis) in an important drop of 93.35\% in the median and 86.96\% in the mean values. Most interesting are the results regarding the state estimation in the y-axis as for the scenario 2 the mean worsens while the median improves. Additionally, concerning state estimation (y-axis) the mean value aggravates while the median improves. This is mainly due to the asymmetry of the corresponding data and their greater variance compared to scenario 1. We conclude that in scenario 2 more entries lie below the median (there is a greater number of accurate estimates), while the distribution of the  data in the upper half is more sparse with most of its values in close proximity to the median or the 75-th percentile. Last, regarding the z-axis, a decrease of 23.45\% in mean and 32.96\% in the median is achieved using all three sensors. 
Concluding, by observing its 25-th and 75-th percentiles we notice a clear reduction in the variance of the state estimation error.
\begin{table}[h]
\begin{center}
\caption{Estimation error (MSE) - ($m$)}
\begin{tabular}{|c|c|c|c|c|c|c|}
\hline
& \multicolumn{3}{c|}{\textbf{IMU\&UWB}} & \multicolumn{3}{c|}{\textbf{IMU\&UWB\&YOLO}} \\
\hline
& \textbf{x} & \textbf{y} & \textbf{z} & \textbf{x} & \textbf{y} & \textbf{z} \\
\hline
\hline
\textbf{mean} & 0.0920 & 0.0071 & 0.0226 & 0.0120 &  0.0412 &  0.0173\\
\hline
\textbf{median} & 0.0993  & 0.0049 & 0.0179 & 0.0066 &  0.0032 & 0.0120 \\
\hline
\textbf{75\%tile} & 0.1210 &  0.0072 & 0.0310 & 0.0082 & 0.0103 &  0.0277\\
\hline
\textbf{25\%tile} & 0.0673 & 0.0041 & 0.0028 & 0.0037 & 0.0020 & 0.0005\\
\hline
\end{tabular}
\label{table:estimation_error}
\end{center}
\end{table}

\noindent Table~\ref{table:path_following_error} depicts the path following MSE of both scenarios considered. Although the path following error is relatively small for scenario 1, there is a significant overall reduction of 35\% in the median MSE of the path following when all sensors (scenario 2) are fused through the Kalman filter.
\begin{table}[htbp]
\begin{center}
\caption{Path following error (MSE) - ($m$)}
\begin{tabular}{|c|c|c|}
\hline
& \textbf{IMU\&UWB} & \textbf{IMU\&UWB\&YOLO} \\
\hline
\hline
\textbf{mean} & 0.0098 & 0.0101 \\
\hline
\textbf{median} & 0.0080  & 0.0052  \\
\hline
\textbf{75\%tile} & 0.0096 &  0.0066 \\
\hline
\textbf{25\%tile} & 0.0071 & 0.0046 \\
\hline
\end{tabular}
\label{table:path_following_error}
\end{center}
\end{table}

\begin{figure}[ht!]
{\parindent-4pt
\begin{tikzpicture}
\begin{axis}
[height=5.4cm,
width=9.1cm,
legend style={font=\footnotesize},
legend columns={-1},
legend cell align={left}, 
smooth,
point meta=explicit,
ylabel shift = 0pt,
minor tick num=1,
ylabel style={yshift=-0.4cm},
axis equal=false,
line width=0.7pt,
no markers,
grid=major,
grid style={dashed,black!50},
        extra x tick style={
        grid=none,
        major tick length=0pt,},
x label style={at={(axis description cs:0.5,0)}},
y label style={at={(axis description cs:0,0.5)}},
 axis on top,
xmin=0,
xmax=20,
ymin=-3.25,
ymax=3.25,
xlabel=position - $x$~(m),ylabel=position - $y$~(m)]
\node[draw=none,anchor=south west,opacity=0.3,inner sep=0pt] at (rel axis cs:-0.05,-0.02) {\includegraphics[scale=0.2]{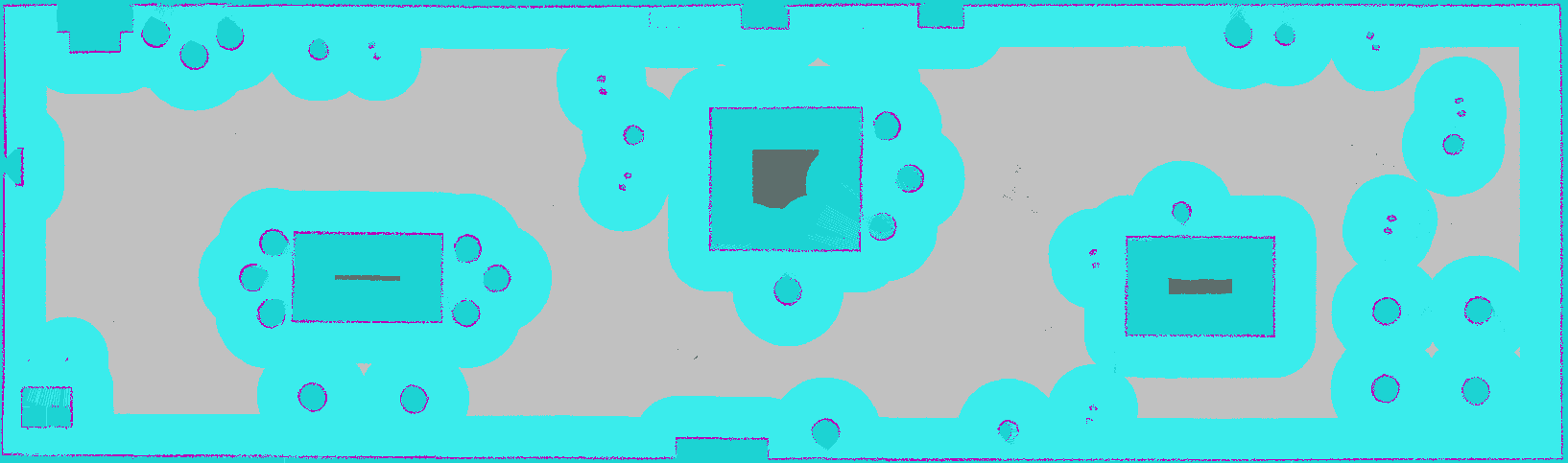}};

\addplot+[dashed, black] table [x=Desired X, y=Desired Y, col sep=comma] {results/imu_uwb_path_following.csv};
\addplot+[blue] table [x=Actual X, y=Actual Y, col sep=comma] {results/imu_uwb_path_following.csv};
\addplot+[red] table [x=Actual X, y=Actual Y, col sep=comma] {results/sensor_fusion_path_following.csv};
\addlegendentry{desired}
\addlegendentry{IMU\&UWB}
\addlegendentry{IMU\&UWB\&YOLO}
\end{axis}
\end{tikzpicture}
\caption{Tracking performance using intermittend multisensor fusion.}
\label{fig:example_of_a_trajectory}}
\end{figure}
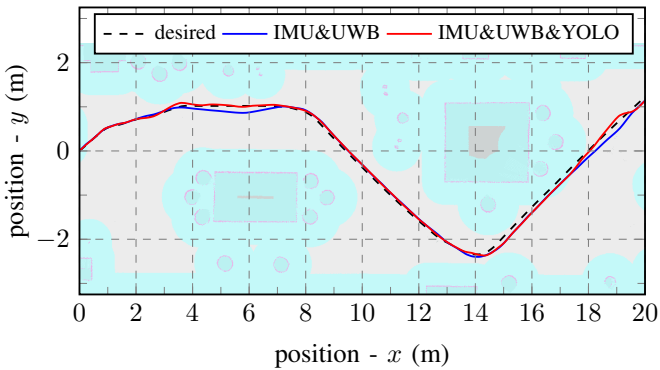

Fig.~\ref{fig:example_of_a_trajectory} illustrates three trajectories along the $xy$-plane, (the reference attitude $z$ is kept constant). The desired planned trajectory is shown in black dashed line, while the actual trajectory of the quadrotor in scenario 1 and scenario 2, are shown in blue and red line, respectively. In some cases, the UWB measurements were deliberately ignored  in order to better validate the contribution of the camera-based localization system. Its contribution is most apparent in $4 \leq x \leq 6m$ range. The resulting trajectories suggest that the indicated setup can handle situations where measurements from all sensors are not continuously available.

%
%
%
%
\section{Conclusions and Future Directions}\label{sec:conclusions}

In this work, we proposed and developed an onboard real-time localization system via sensor fusion that handles sensor measurements arriving intermittently at the estimator, due to unreliable communication channels and/or incompetence of sensors to provide measurements. To evaluate the system's performance, we designed a robust hybrid multisensory fusion framework for autonomous on-board indoor navigation, and a framework for HIL simulation of autonomous on-board indoor navigation. This is the first work, to the best of the authors' knowledge, that fuses such noisy sensor measurements with intermittent communication for onboard quadrotor pose estimation and navigation. 

Part of ongoing work focuses on anticipating the communication and computation delays invoked, in order to improve the robustness of the overall system. We are implementing the whole experimental setup in order to see how our approach behaves in realistic scenarios.
Additionally, performance evaluation could be performed in dynamically changing environments. 


\bibliography{paper}
\bibliographystyle{IEEEtran}

\end{document}